\documentclass[10pt,twocolumn,letterpaper]{article}

\usepackage{cvpr}
\usepackage{times}
\usepackage{epsfig}
\usepackage{graphicx}
\usepackage{amsmath}
\usepackage{amssymb}
\usepackage{comment}
\usepackage{subfigure}
\usepackage{pifont}%

\usepackage{booktabs}
\usepackage{array, makecell, tabularx}
\usepackage{xspace}

\usepackage{subfigure}
\usepackage[dvipsnames]{xcolor}
\usepackage{rotating}
\usepackage[export]{adjustbox}

\usepackage[pagebackref=true,breaklinks=true,letterpaper=true,colorlinks,bookmarks=false]{hyperref}

\newcommand{\approxtilde}{{\raise.17ex\hbox{$\scriptstyle\sim$}}}

\makeatletter
\def\@rothead[#1]#2{\thead{\\[-.65\normalbaselineskip]
  \turn{\cellrotangle}\thead[#1]{#2}\endturn}}
\makeatother

\cvprfinalcopy %

\def\cvprPaperID{4201} %

\ifcvprfinal\fi

\newcommand{\yes}{\textcolor{ForestGreen}{\checkmark}}  %
\newcommand{\yesyes}{\textcolor{ForestGreen}{\checkmark$^+$}}  %
\newcommand{\yesno}{\textcolor{ForestGreen}{\checkmark}\textcolor{red}{$^-$}}  %
\newcommand{\no}{\textcolor{red}{\ding{55}}}  %

\begin{document}

 \author{
     Jonathan Tremblay\\
 		NVIDIA \\
     {\tt\small jtremblay@nvidia.com}
     \and
     Thang To\\
 		NVIDIA \\
     {\tt\small thangt@nvidia.com}
     \and
     Stan Birchfield\\
 		NVIDIA \\
     {\tt\small sbirchfield@nvidia.com}
 }

\title{Falling Things:  A Synthetic Dataset for \\ 3D Object Detection and Pose Estimation}
\maketitle
\thispagestyle{empty}

\begin{abstract}

We present a new dataset, called Falling Things (FAT), for advancing the state-of-the-art in object detection and 3D pose estimation in the context of robotics.\footnote{The dataset can be downloaded from \url{http://research.nvidia.com/publication/2018-06_Falling-Things}} 
By synthetically combining object models and backgrounds of complex
composition and high graphical quality, we are able to generate photorealistic images with accurate 3D pose annotations for all objects in all images. 
Our dataset contains 
60k 
annotated photos of 21 household objects taken from 
the YCB dataset \cite{calli2015icar:ycb}. 
For each image, we provide the 3D poses, 
per-pixel class segmentation, and 2D/3D bounding box coordinates for all objects.
To facilitate testing different input modalities, we provide mono and stereo RGB images, along with registered dense depth images.  
We describe in detail the generation process and 
statistical analysis of the data.

\end{abstract}
\section{Introduction}
\label{sec:introduction}

Robotic manipulation of household objects in everyday environments requires accurate detection and pose estimation of multiple object categories.  
This presents a two-fold challenge for developing robotic perception algorithms.  
First, acquiring ground truth data is time-consuming, 
error-prone, and potentially expensive (depending upon the technique used).
This limits the {\em evaluation} of algorithms, particularly with respect to new object categories or environmental conditions.
Secondly, existing techniques for acquiring real world 
data \cite{xiang2017arx:posecnn,hinterstoisser2012accv:linemod,xiang2014wacv} do not scale.
As a result, they are not capable of generating 
the large datasets that are needed for training deep neural networks.

\begin{figure}[h!]
    \centering
    \begin{tabular}{ccc}
        \multicolumn{3}{c}{
            \includegraphics[width=0.95\columnwidth]
                {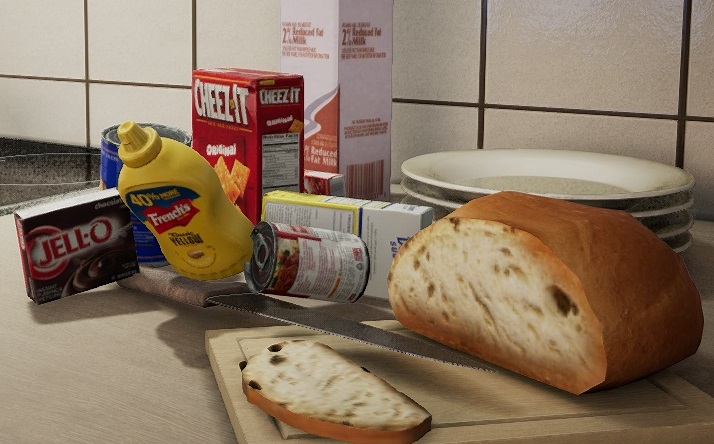}
        }\\
        \includegraphics[width=0.2825\columnwidth]
            {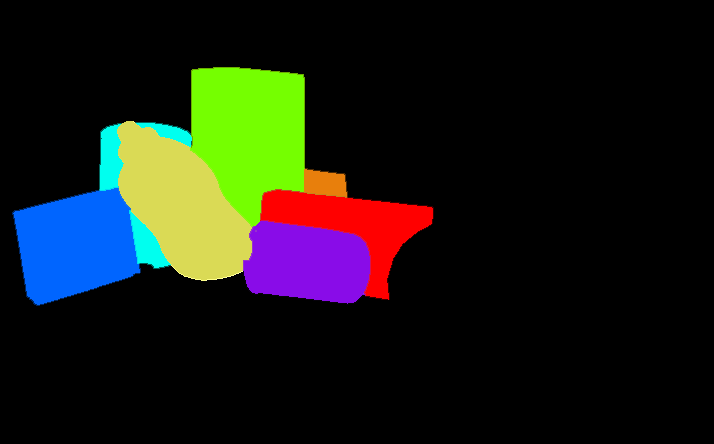} &
        \includegraphics[width=0.2825\columnwidth]
            {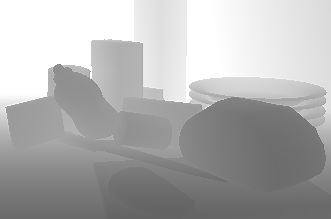} &
        \includegraphics[width=0.2825\columnwidth]
            {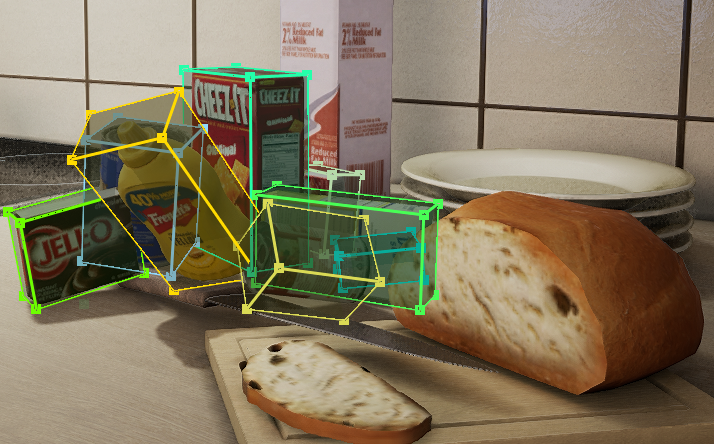}
%
    \end{tabular}
    \vspace{-1mm}
    \vspace{-1mm}
    \caption{The Falling Things (FAT) dataset was generated by placing 3D household object models ({\em e.g.}, mustard bottle, soup can, gelatin box, etc.) in virtual environments.  Each snapshot consists of a stereo pair of RGB images (only one of which is shown, top), pixelwise segmentation of the objects (bottom left), depth (bottom center), 2D/3D bounding box coordinates (bottom right), and 3D poses of all objects (not shown).}
    \label{fig:example}
    \vspace{-3mm}
\end{figure}

We propose to overcome such limitations by using synthetically generated data.  
Synthetic data has been gaining traction in recent years as an efficient means of both training and evaluating DNNs for computer vision problems for which collecting ground truth data is laborious, {\em e.g.}, 
stereo \cite{zhang2016arx:unst}, 
optical/scene flow \cite{mayer2015arx:flythings}, 
or semantic segmentation \cite{gaidon2016arx:vkitti}.  
We believe that 3D detection and pose estimation fall within this category and are thus a natural fit for synthetic data due to the difficulty of acquiring accurate ground truth.

In this paper we introduce the \emph{\underline{Fa}lling \underline{T}hings (FAT)} dataset, consisting of a large number (61,500) of 
snapshots for training and evaluating robotics scene understanding algorithms in household environments.  
Specifically, as shown in Fig.~\ref{fig:example}, each snapshot consists of a stereo pair of image frames with corresponding depth images, along with the 3D poses, 2D/3D bounding boxes, projected 3D bounding boxes, and pixelwise segmentation of the known objects in the scene.  
Unlike previous datasets, we aim for photorealistic images of real-world objects, leveraging the popular YCB~object models \cite{calli2015icar:ycb} placed in different virtual environments.

To better understand the contribution of our dataset, consult Table~\ref{table:comparison}.  To our knowledge, only two datasets exist with accurate ground truth poses of multiple objects with significant occlusion (T-LESS \cite{hodan2017wacv:tless}, YCB-Video \cite{xiang2017arx:posecnn}).  
Yet neither of these datasets contain extreme lighting variations or multiple modalities.  Our FAT dataset thus extends upon these existing solutions in both quantity and variety.

\begin{table*}
\begin{center}
\begin{footnotesize}
\vspace{-10mm}
\begin{tabular}{rrrccccccccccc}

Dataset 
& \# objects 
& \# frames
& description
& \rotatebox{75}{depth}
& \rotatebox{75}{stereo}
& \rotatebox{75}{3D pose}
& \rotatebox{75}{full rotation}
& \rotatebox{75}{occlusion}
& \rotatebox{75}{extreme lighting}
& \rotatebox{75}{segmentation}
& \rotatebox{75}{bbox coords} \\
\midrule
   EPFL multi-view \cite{ozuysal2009cvpr:pose} 
        & 20 & 2k & cars & \no & \no & \yesno & \no & \no & \no & \no & \no \\
   UW RGBD \cite{lai2011icra:rgbd} & 300 & 250k & household & \yes & \no & \yesno & \yes & \no & \no & \no & \no \\
   LINEMOD \cite{hinterstoisser2012accv:linemod} 
        & 15 & 18k & household & \yes & \no & \hspace{-1.8ex}\yes & \yes & \no & \no & \no & \no \\
   Object tracking \cite{choi2013iros:objt}
     & 4 & 6k & household & \yes & \no & \hspace{-1.8ex}\yes & \yes & \no & \no & \no & \no \\
     Pascal3D+ \cite{xiang2014wacv} & 12 & 30k & various & \no & \no & \hspace{-1.8ex}\yes & \no & \yes & \yes & \no & \no \\
		Brachmann et al.~\cite{brachmann2014eccv:occlusion} & 20 & 10k & various & \yes & \no & \hspace{-1.8ex}\yes & \yes & \yes & \yes & \no & \no \\
   Occlusion \cite{brachmann2014eccv:occlusion,krull2015iccv:rgbd} 
        & 8 & 1k & household & \yes & \no & \hspace{-1.8ex}\yes & \yes & \yes & \no & \no & \no \\
Krull et al.~\cite{krull2014accv:6dof}
	& 3 & 3k & handheld & \yes & \no & \hspace{-1.8ex}\yes & \yes & \yes & \no & \no & \no \\
   Rutgers APC \cite{rennie2016ral:rgbd} 
            & 24 & 10k & warehouse & \yes & \no & \hspace{-1.8ex}\yes & \no & \no & \no & \no & \no \\
   T-LESS \cite{hodan2017wacv:tless} 
       & 30 & 10k & industrial & \yes & \no & \yesyes & \yes & \yes & \no & \no & \no \\
   YCB-Video \cite{xiang2017arx:posecnn} 
        & 21 & 134k & household & \yes & \no & \yesyes & \yes & \yes & \no & \yes & \yes \\
     FAT (ours) 
            & 21 & 60k & household & \yes & \yes & \yesyes & \yes & \yes & \yes & \yes & \yes \\
\bottomrule
\end{tabular}
\end{footnotesize}
\caption{Datasets for object detection and pose estimation.  Under 3D pose, \yesyes means that the pose of all known objects in the scene are provided, \yes means only the pose of a single object is provided, and \yesno means that the provided poses are approximate.  Note that segmentation and bounding boxes can be determined for any dataset by overlaying models according to ground truth.}
\label{table:comparison}
\end{center}
\end{table*}

\section{Falling Things Dataset}
\label{sec:method}

\noindent\textbf{Description.}\hspace{1mm} 
All data were generated by a custom plugin we developed for Unreal Engine 4 (UE4) \cite{to2018ndds}.
By leveraging asynchronous, multithreaded sequential frame grabbing, the plugin generates data 
at a rate of 50--100~Hz, which is significantly faster than either the default UE4 screenshot function or the publicly available UnrealCV tool \cite{qiu2016arx:uncv}.

We selected three virtual environments within UE4:  a kitchen, sun temple, and forest. 
These environments were chosen for their high-fidelity modeling and quality, as well as for the variety of indoor and outdoor scenes. 
For each environment we manually selected five specific locations covering a variety of terrain and lighting conditions ({\em e.g.}, on a kitchen counter or tile floor, next to a rock, above a grassy field, and so forth).  Together, these yielded 15 different locations consisting of a variety of 3D backgrounds, lighting conditions, and shadows.

We selected 21 household objects\footnote{We used the same subset of 21 objects as in \cite{xiang2017arx:posecnn}.} from the YCB \cite{calli2015icar:ycb} dataset.\footnote{\url{http://www.ycbbenchmarks.com/object-models}}  
Since these models are not all aligned, translation was applied to each of the downloaded models to center the coordinate frame at the object centroid, and rotation was applied to align the coordinate axes with those of the object, taking product labeling into account.  
For each run, some of these object models were placed at random positions and orientations within a vertical cylinder of radius $5$~cm and height of $10$~cm placed at a fixation point.
The objects' initial positions are sampled within this volume to avoid initial penetration.
The objects were then allowed to fall under the force of gravity, as well as to collide with one another and with the surfaces in the scene.  
While the objects fell, the virtual camera system was rapidly teleported to random azimuths, elevations, and distances with respect to the fixation point to collect data.
Azimuth ranged from --$120^\circ$ to +$120^\circ$ (to avoid collision with the wall, when present), elevation from $5^\circ$ to $85^\circ$, and distance from 0.5~m to 1.5~m.  

Our virtual camera system consists of a pair of stereo RGBD cameras.  
This design decision allows the dataset to support at least three different sensor modalities.
Whereas single RGBD sensors are commonly used within robotics, stereo sensors have the potential to yield higher quality output with fewer distortions, and a monocular RGB camera has obvious advantages in terms of cost, simplicity, and availability.  By supporting all of these options, the dataset allows researchers to use their modality of choice, as well as to explore the important topic of comparing across modalities.
In our system, the baseline separating the left and right cameras is 6~cm; and each camera's horizontal field of view is 64 degrees, leading to a focal length of 768.2 pixels with an image resolution of $960 \times 540$.

The dataset (consisting of 61,500 unique frames) is divided into two parts:\\
\vspace{-4ex}
\begin{itemize}
\setlength{\itemsep}{-3.5ex}
	\item 
\textit{Single objects.}\hspace{1mm} 
The first part of the dataset was generated by dropping each object model in isolation
$\sim$5 times at each of the 15 locations. 
For each run, $\sim$20 image frames were generated by taking screenshots at a rate of 10~Hz while the object fell for $\sim$2~s. 
Therefore, this part of the dataset consists of 1500 ($100 \times 15$) unique images 
for each of the 21 objects, thus totaling 31,500 frames. 
\\
	\item 
\textit{Mixed objects.}\hspace{1mm} 
The second part of the dataset was generated in the same manner except with a random number of objects sampled uniformly from 2 to 10. 
By sampling the objects with replacement, we allow multiple instances of the same category in an image, unlike many previous datasets.
For each location we generated 2,000 images, thus yielding 30,000 frames.
\end{itemize}
\vspace{-2ex}

\vspace{1.2ex}	
\noindent\textbf{Testing.}\hspace{1mm}
There are several natural ways to split the FAT dataset for training and testing.  
One approach would be to hold out one location per scene as the test sets, and leave the other data for training.  
Another approach would be to hold out one environment for testing, leaving the others for training.
Finally, the single object images could be used for training, while using the remaining images for testing---that is, assuming that occlusion is artificially introduced in the data augmentation process during training.
Further details regarding training/testing methodology can be found with the dataset.

\begin{figure}
    \includegraphics[width=0.9\columnwidth]
        {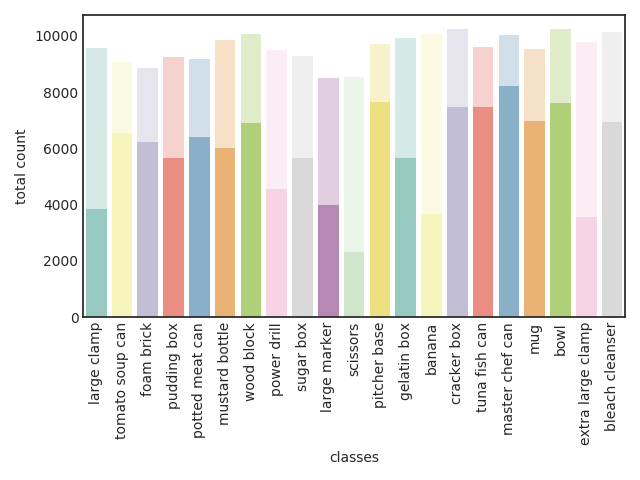}
    \caption{Total appearance count of the 21 YCB objects in the FAT dataset.   Light color bars indicate object visibility greater than 25\%, while solid bars indicate visibility greater than 75\%. (100\% visible means \emph{not occluded}, 0\% visible means \emph{fully occluded}.)}
    \label{fig:dist}
    \vspace{-3mm}

\end{figure}

\vspace{1.2ex}	
\noindent\textbf{Statistics.}\hspace{1mm}
Fig.~\ref{fig:dist} displays the total number of occurrences of each object class in the FAT dataset, using opacity of the color bars to indicate the percentage of the object that is visible (i.e., non-occluded). 
That is, solid color bars show the number of occurrences for which the object is at least 75\% visible, whereas lighter color bars indicate the occurrences for which the visibility is between 25\% and 75\%. (Occurrences with visibility less than 25\% are not shown.) As can be seen, smaller objects (such as the scissors or marker) are occluded more often than larger objects (such as the cracker box).

\begin{figure}
\centering
    \begin{tabular}{cc}
        \includegraphics[width=0.45\columnwidth]
            {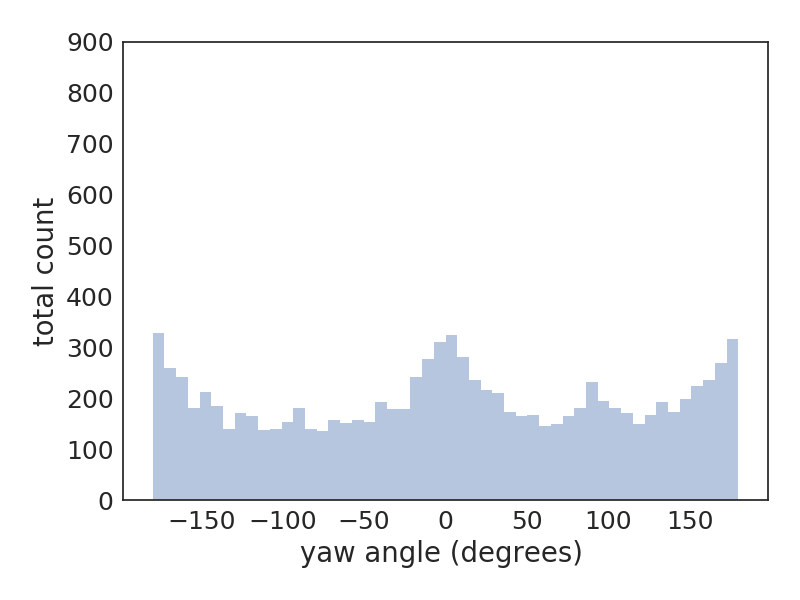} &
        \includegraphics[width=0.45\columnwidth]
            {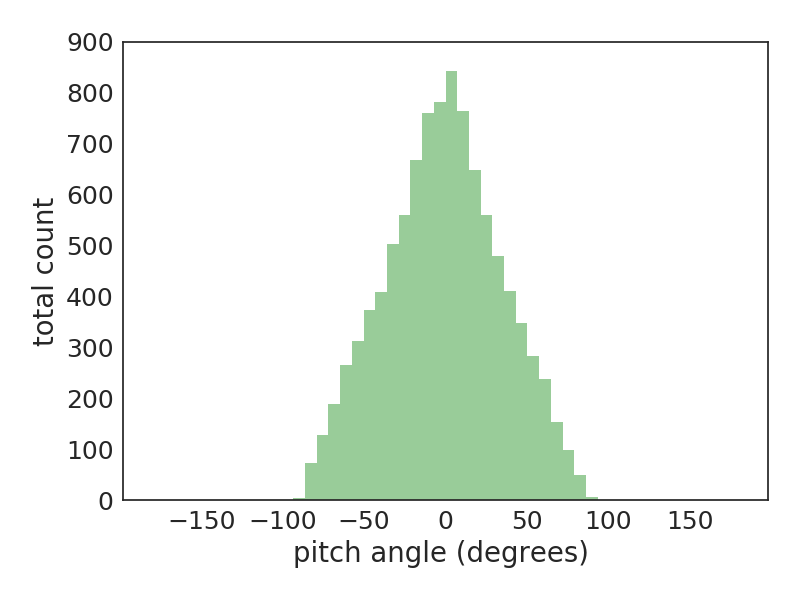} \\
        \includegraphics[width=0.45\columnwidth]
            {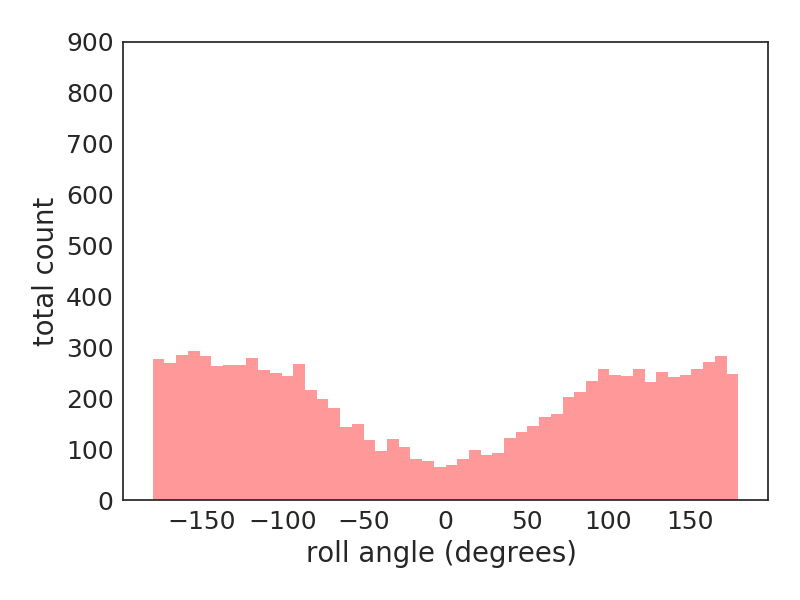} &
        \includegraphics[width=0.45\columnwidth]
            {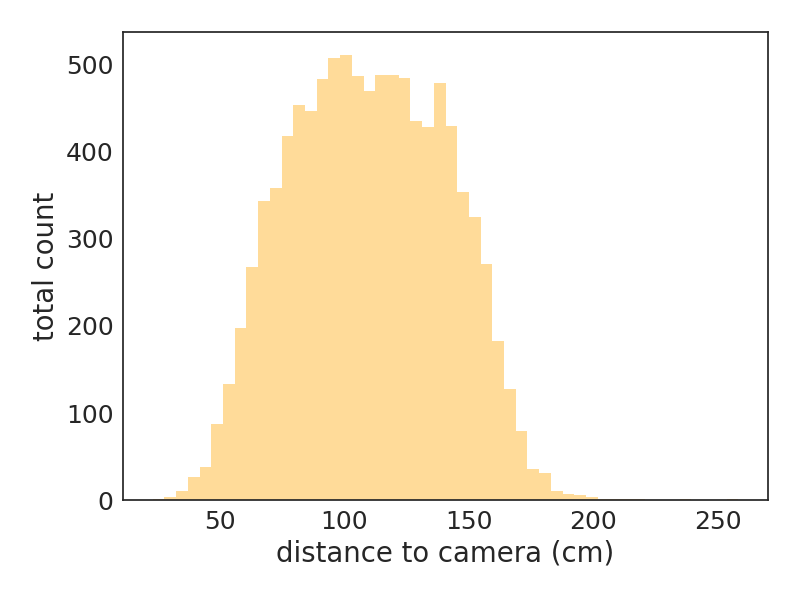} \\ 
        \includegraphics[width=0.45\columnwidth]
            {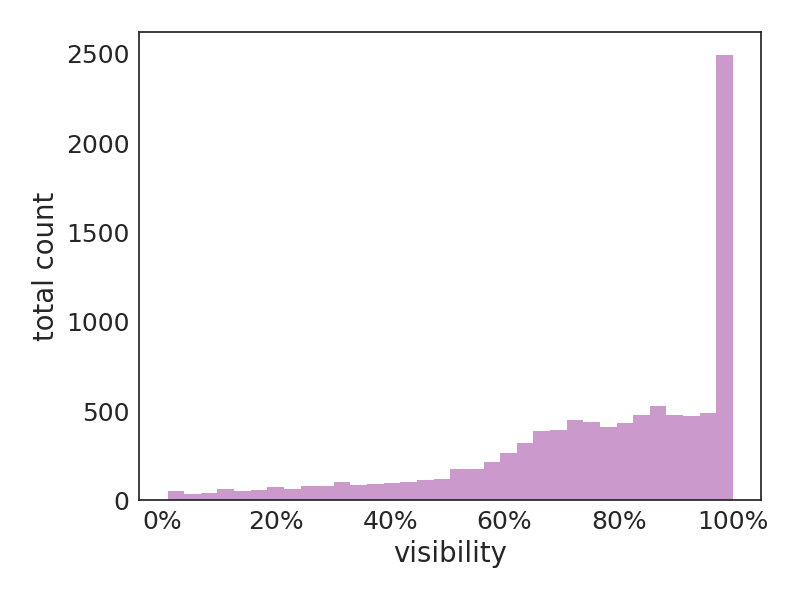} & 
        \includegraphics[width=0.45\columnwidth]
            {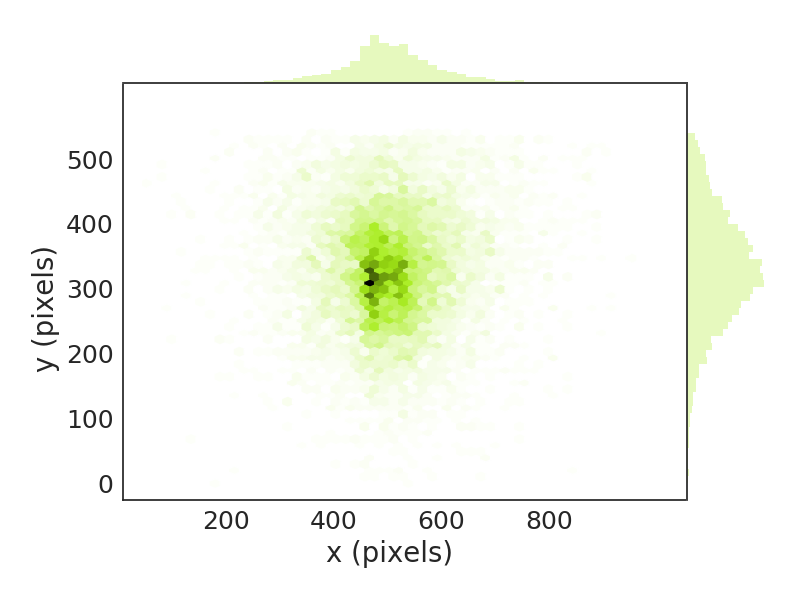}              
            \vspace{-1mm}
    \end{tabular}
    \caption{Statistics for one object (mustard bottle) in the FAT dataset. 
    In lexicographic order are the distribution of yaw, pitch, and roll angles; distance to the camera; 
    visibility; and position of the object's centroid within the exported images.}
    \label{fig:stats_mustard}
\end{figure}

Fig.~\ref{fig:stats_mustard} shows statistical distributions of various parameters of one of the YCB objects (the mustard bottle) in 
the FAT dataset (from left camera perspective). 
Shown are the yaw, pitch, and roll of the object with respect to the front.   
The modes in yaw (at 0$^\circ$ and 180$^\circ$ and, to a lesser extent, at $\pm 90^\circ$) are due to the fact that after falling, the object likely lies on either its front or back side.  
Similarly, the angle of the camera with respect to the resting surface biases the pitch angle toward neutrality.  
In contrast, the roll is uniform.

Also shown in the figure is the distribution of the distance to the camera, 
which extends slightly beyond the intended range of 0.5 m to 1.5 m because oftentimes objects roll or slide after impact. 
The final row shows the visibility indicating that, while the single objects are fully visible, significant occlusions occur; and the distribution of the centroid within the image, which is approximately a broad Gaussian centered near the center of the image.
Similar distributions were observed for other YCB objects.

\section{Conclusion}
\label{sec:conclusion}

We have presented a new dataset to accelerate research in  
object detection and pose estimation, 
as well as segmentation, depth estimation, and sensor modalities. 
The proposed dataset focuses on household items from the YCB dataset and has been rendered with high fidelity and a wide variety of backgrounds, poses, occlusions, and lighting conditions. 
Statistics from the dataset confirm this variety quantitatively.
We hope that researchers find this dataset useful for exploring 
robust solutions to open problems 
such as object detection, pose estimation, depth estimation from monocular and/or stereo cameras,
and depth-based segmentation, to advance the field of robotic manipulation.

{\small
\bibliographystyle{ieee}
\bibliography{main}
}

\clearpage
\newpage

\onecolumn 
\appendix
\section{Sample Images}

The figure below shows sample images from the FAT dataset, demonstrating the variety of object poses, backgrounds, composition, and lighting conditions.  (Random center crops are shown.)

\begin{center}
    \begin{tabular}{c}
        \includegraphics[width=0.95\columnwidth]
            {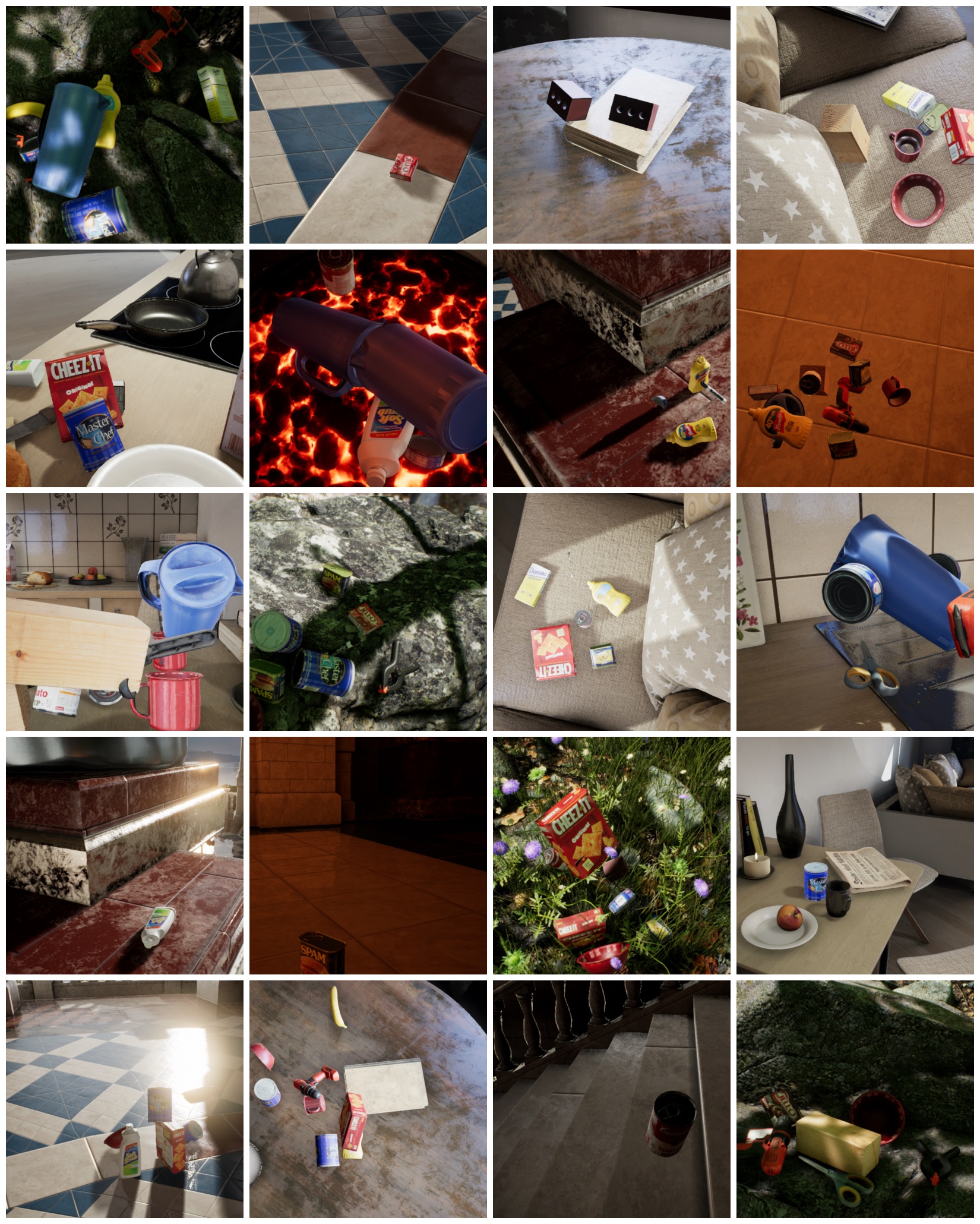}              
            \vspace{-1mm}
    \end{tabular}
\end{center}

\pagebreak

\clearpage
\newpage

\onecolumn 
\section{YCB Objects}

The figure below shows the 21 YCB objects used in creating the dataset:
pitcher base (019), bleach cleanser (021), cracker box (003), power drill (035), wood block (036),
mustard bottle (006), sugar box (004), 
potted meat can (010), mug (025), tomato soup can (005), large marker (040), master chef can (002), foam brick (061), pudding box (008), 
bowl (024), banana (011), large clamp (051), scissors (037), tuna fish can (007), gelatin box (009), and extra large clamp (052).  Each number in parentheses indicates the object's YCB number.\footnote{See \url{http://ycb-benchmarks.s3-website-us-east-1.amazonaws.com/}}

\begin{center}
    \begin{tabular}{c}
        \includegraphics[width=0.95\columnwidth]
            {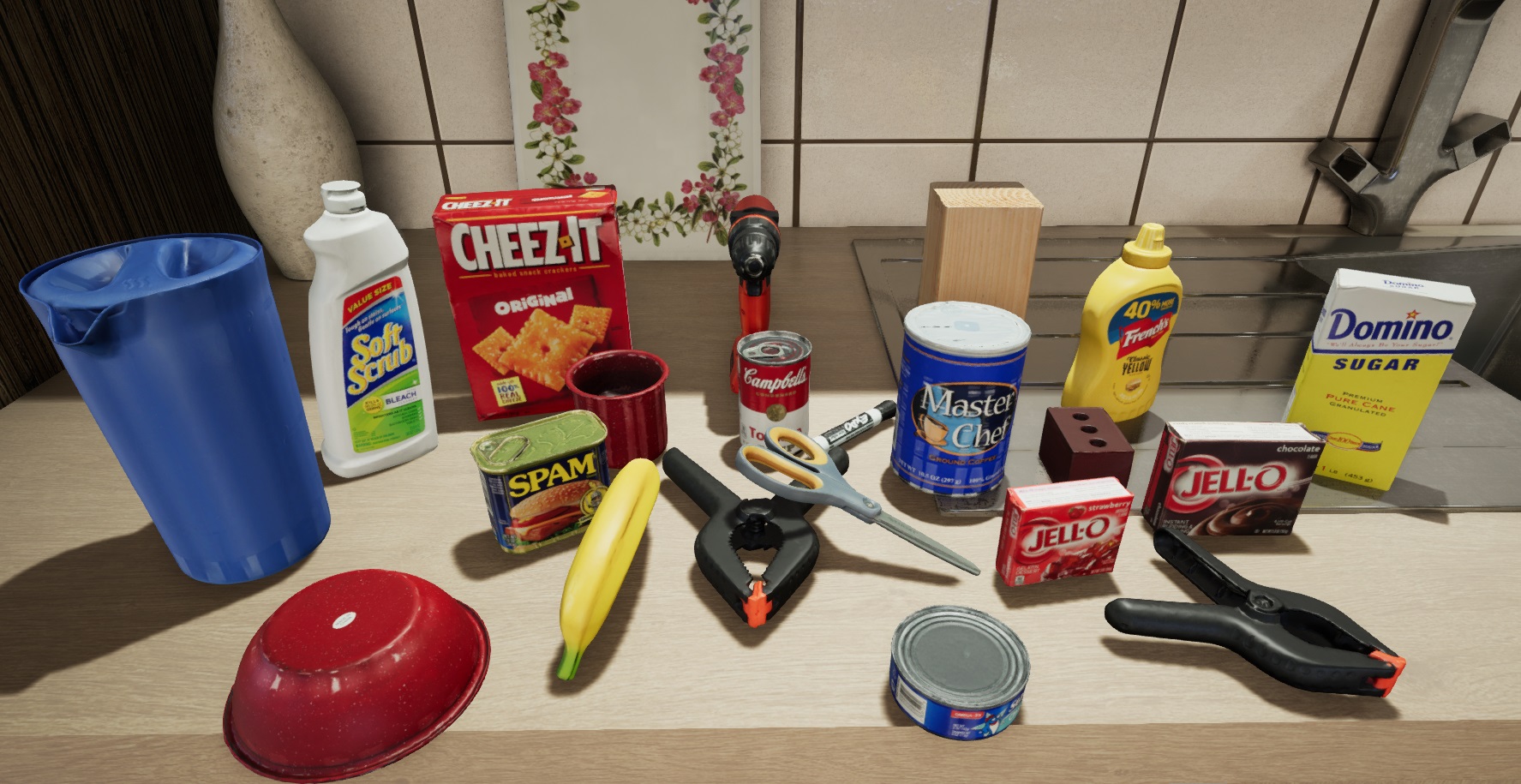}              
            \vspace{-1mm}
    \end{tabular}
\end{center}

\end{document}